\documentclass[journal,twoside,web]{ieeecolor}
\usepackage{generic}
\usepackage{cite}
\usepackage{amsmath,amssymb,amsfonts}
\usepackage{algorithmic}
\usepackage{graphicx}
\usepackage{algorithm,algorithmic}
\usepackage{hyperref}
\hypersetup{hidelinks=true}
\usepackage{textcomp}

\usepackage{multirow}
\usepackage{makecell}

\usepackage{threeparttable}
\usepackage{comment}

\def\BibTeX{{\rm B\kern-.05em{\sc i\kern-.025em b}\kern-.08em
    T\kern-.1667em\lower.7ex\hbox{E}\kern-.125emX}}
\begin{document}
\title{Transformer-Based Multi-Region Segmentation and Radiomic Analysis of HR-pQCT Imaging for Osteoporosis Classification}
\author{Mohseu Rashid Subah, \IEEEmembership{Member, IEEE}, Mohammed Abdul Gani Zilani, Thomas L. Nickolas, Matthew R. Allen, Stuart J. Warden, and Rachel K. Surowiec, \IEEEmembership{Member, IEEE}
\thanks{This work is supported by the National Institutes of Health (NIH/NIAMS P30 AR072581 and LRP 1L30DK130133-0) and a Norman S. Coplon Extramural Grant. This work involved human subjects in its research. Approval of all ethical and experimental procedures and protocols was granted by the Institutional Review Boards of Indiana University (IRB protocol \#1707550885, date of approval 11/05/2025) and Columbia University Irving Medical Center (IRB protocol \#AAAM7850, date of approval 02/04/2014), and the study was performed in line with the Declaration of Helsinki.}
\thanks{Mohseu Rashid Subah is with the Weldon School of Biomedical Engineering, Purdue University, West Lafayette, IN, USA (e-mail: msubah@purdue.edu) and holds an AAUW International Fellowship.}
\thanks{Mohammed Abdul Gani Zilani is with the Weldon School of Biomedical Engineering, Purdue University, West Lafayette, IN, USA (e-mail: mzilani@purdue.edu).}
\thanks{Thomas L. Nickolas is with the Division of Bone and Mineral Diseases, Washington University Medicine, St. Louis, MO, USA (e-mail: thomasn@wustl.edu).}
\thanks{Matthew R. Allen is with the Department of Anatomy, Cell Biology \& Physiology, Indiana University School of Medicine, Indianapolis, IN, USA (e-mail: Matallen@iu.edu).}
\thanks{Stuart J. Warden is with the Department of Physical Therapy, School of Health \& Human Sciences, Indiana University Indianapolis, Indianapolis, IN, USA (e-mail: stwarden@iu.edu).}
\thanks{Rachel Surowiec is with the Weldon School of Biomedical Engineering, Purdue University, West Lafayette, IN, USA (e-mail: rsurowie@purdue.edu).}}

\maketitle

\begin{abstract}
Osteoporosis is a skeletal disease typically diagnosed using dual-energy X-ray absorptiometry (DXA), which quantifies areal bone mineral density but overlooks bone microarchitecture and surrounding soft tissues. High-resolution peripheral quantitative computed tomography (HR-pQCT) enables three-dimensional microstructural imaging with minimal radiation. However, current analysis pipelines largely focus on mineralized bone compartments, leaving much of the acquired image data underutilized. We introduce a fully automated framework for binary osteoporosis classification using radiomics features extracted from anatomically segmented HR-pQCT images. To our knowledge, this work is the first to leverage a transformer-based segmentation architecture, i.e., the SegFormer, for fully automated multi-region HR-pQCT analysis. The SegFormer model simultaneously delineated the cortical and trabecular bone of the tibia and fibula along with surrounding soft tissues and achieved a mean F1 score of 95.36\%. Soft tissues were further subdivided into skin, myotendinous, and adipose regions through post-processing. From each region, 939 radiomic features were extracted and dimensionally reduced to train six machine learning classifiers on an independent dataset comprising 20,496 images from 122 HR-pQCT scans. The best image level performance was achieved using myotendinous tissue features, yielding an accuracy of 80.08\% and an area under the receiver operating characteristic curve (AUROC) of 0.85, outperforming bone-based models. At the patient level, replacing standard biological, DXA, and HR-pQCT parameters with soft tissue radiomics improved AUROC from 0.792 to 0.875. These findings demonstrate that automated, multi-region HR-pQCT segmentation enables the extraction of clinically informative signals beyond bone alone, highlighting the importance of integrated tissue assessment for enhanced osteoporosis detection. 
\end{abstract}


\section{Introduction}
\label{sec:introduction}
\IEEEPARstart{O}{steoporosis} is the most prevalent bone disease worldwide that results in reduced bone mineral density (BMD), thereby predisposing individuals to fragility fractures from low-energy trauma \cite{Kanis2000, Melton1998}. Because the disease often remains undetected until a fracture occurs, delayed diagnosis can lead to severe complications, long-term disability, and increased mortality, underscoring the need for timely diagnosis \cite{Compston2019, Curtis2017}. Clinically, osteoporosis is diagnosed using dual-energy X-ray absorptiometry (DXA), with an areal bone mineral density (aBMD) T-score below $-2.5$ indicating disease \cite{Kanis2002}. However, osteoporosis represents a systemic musculoskeletal disorder that extends beyond mineral density alone, in which changes in bone composition and quality coexist with alterations in surrounding soft tissues \cite{Bogl2010, Reid2002}. 

Prior studies have shown that osteoporosis is characterized by early deterioration of trabecular bone, followed by progressive loss of cortical bone with aging, highlighting region-specific disease processes \cite{Osterhoff2016}. Sarcopenia, a disease characterized by loss of muscle strength, mass, and density with increased intramuscular fat infiltration \cite{Warden2025}, is frequently associated with osteoporosis \cite{Laurent2019}. Consistent with this, muscle-based measures, such as the psoas muscle index, have been shown to correlate significantly with BMD and osteoporotic fracture risk \cite{Kajiki2022}. Despite its widespread availability and low radiation dose, DXA cannot capture three-dimensional volumes and relies solely on bone mineral content, preventing assessment of bone size, geometry, or microarchitecture. Moreover, DXA lacks the resolution to evaluate soft tissue quality or distribution, thereby limiting its ability to capture tissue-specific changes associated with osteoporosis \cite{Williams2021, Choksi2018}.

High-resolution peripheral quantitative computed tomography (HR-pQCT) offers a detailed assessment ($<60.7$ µm nominal isotropic voxel size) of peripheral bones \cite{Fuller2015} and adjacent soft tissues \cite{Warden2025} using minimal radiation ($<5$ µSv). It provides a variety of structural and density parameters and allows the estimation of bone mechanical properties by employing finite element analysis (FEA) modeling \cite{Fuller2015}. Several studies have demonstrated that HR-pQCT metrics improve osteoporotic fracture prediction beyond DXA measures \cite{Fink2018, SornayRendu2017, Burt2018, Samelson2019}. Furthermore, HR-pQCT has been used to characterize specific osteoporosis subgroups, such as lactation-associated osteoporosis \cite{Agarwal2024}, male osteoporosis \cite{Okazaki2016}, and osteosarcopenia \cite{Cheng2024}. However, most prior work has examined the association between conventional HR-pQCT parameters and low BMD, while advanced computational methods such as machine learning, deep learning, and radiomics remain underexplored for distinguishing osteoporosis using this modality. 

Radiomics is a technique that extracts numerous quantitative features characterizing the texture and intensity distribution in medical images, capturing information not visually discernible \cite{Mayerhoefer2020}. Several studies have explored radiomics for osteoporosis detection, typically defining the disease using BMD derived from DXA or quantitative CT (QCT) and performing either binary (osteoporosis vs. normal) or multi-class classification (osteoporosis, osteopenia, and normal). These studies have used imaging modalities such as hip and panoramic radiographs \cite{Kim2022, Fanelli2025}, abdominal, lumbar spine, or low-dose chest CT \cite{Wang2023, Jiang2022, Huang2022, Xue2022, Tong2024}, QCT \cite{Xie2022}, and lumbar spine magnetic resonance imaging (MRI) \cite{He2021}, demonstrating the robustness of radiomics features across diverse imaging domains. While most investigations focused on bone regions such as the femur or vertebrae, Huang \textit{et al.} \cite{Huang2022} demonstrated that radiomics features derived from the psoas muscle on abdominal CT, when combined with a gradient boosting model, can be used to predict osteoporosis. 

However, existing studies largely examine a single anatomical region or tissue type and rarely leverage high-resolution modalities capable of separating cortical, trabecular, and surrounding soft-tissue compartments. As a result, the comparative diagnostic value of radiomics features from distinct compartments remains poorly understood. The high spatial resolution of HR-pQCT makes it well suited for radiomics analysis, as fine-scale texture and intensity patterns relevant to tissue microstructure can be captured more sensitively than with conventional clinical imaging. Therefore, we hypothesize that radiomics features derived from anatomically and functionally distinct regions of HR-pQCT scans capture complementary aspects of osteoporosis, with soft tissue features providing diagnostic value beyond traditional bone-centric approaches. 

Given the growing clinical and scientific interest in jointly characterizing bone and muscle health, automated segmentation tools capable of delineating distinct anatomical regions are essential for enabling reproducible, large-scale data analysis. Existing HR-pQCT bone segmentation efforts primarily relied on image processing-based methods \cite{Valentinitsch2012, Hafri2016, Ohs2021, Klintstrm2024, Zhou2025}. Although these methods have advanced HR-pQCT segmentation, the current gold standard protocol \cite{Burghardt2010} remains semi-automatic and requires manual correction, which is time-consuming, labor-intensive, and prone to intra- and inter-operator variability, resulting in precision loss \cite{Whittier2020}. Neeteson \textit{et al.} \cite{Neeteson2023} introduced the first fully automatic method for segmenting cortical and trabecular compartments of the radius or tibia in HR-pQCT images using a modified U-Net architecture. However, convolutional neural networks (CNNs) such as the U-Net \cite{Ronneberger2015} inherently struggle with modeling long-range dependencies, often limiting their ability to capture complex spatial relationships in high-resolution medical images. Vision transformers \cite{ViT} address this limitation by leveraging self-attention mechanisms to model global context while maintaining local precision. We hypothesize that a transformer-based, fully automated multi-region segmentation framework, such as the SegFormer \cite{SegFormer}, can robustly and reproducibly delineate both mineralized and soft-tissue compartments in HR-pQCT images, enabling downstream radiomics analysis.

In summary, the main contributions of this work are as follows:  

\begin{itemize}

\item We meticulously annotate and introduce the first HR-pQCT segmentation dataset comprising 6,720 distal tibia images with corresponding five-class pixel-wise ground truth masks. 

\item We leverage a transformer-based segmentation network, i.e., the SegFormer, to segment HR-pQCT images into five distinct regions: the cortical and trabecular compartments of both tibia and fibula, and soft tissues. To our knowledge, this is the first deep learning-based approach to achieve this level of multi-region HR-pQCT segmentation in a fully automatic, end-to-end manner. 

\item We implement a post-processing pipeline to further subdivide the soft tissue region into skin, myotendinous tissue, and adipose tissue. This yields a complete seven-class segmentation when combined with the deep learning output, as shown in Fig. \ref{fig: segmentation soft tissue}.

\item We extract and analyze radiomics features from the segmented regions to classify osteoporosis from a single two-dimensional (2D) image using six machine learning classifiers. To the best of our knowledge, this is the first systematic analysis comparing the diagnostic contributions of distinct bone compartments and soft tissue types for binary osteoporosis classification. 

\item We develop a multivariate logistic regression framework integrating biological variables, physical function measures, DXA-derived mass features, and standard HR pQCT parameters to evaluate their combined predictive value relative to radiomics features for patient level osteoporosis diagnosis.  

\end{itemize}

Overall, this work provides a comprehensive framework integrating deep learning-based segmentation, radiomics, and machine learning for binary osteoporosis classification using HR-pQCT. A preliminary version of this work has been previously reported in \cite{Subah2024, Subah2025}.

\section{Materials and Methods}
\subsection{Study Population and Data Acquisition} In this study, two distinct datasets were used for the segmentation and osteoporosis prediction tasks. Written informed consent was obtained from all participants. All study procedures were approved by the Institutional Review Boards of Indiana University (IRB protocol \#1707550885, date of approval 11/05/2025) and Columbia University Irving Medical Center (IRB protocol \#AAAM7850, date of approval 02/04/2014) and were conducted in accordance with applicable guidelines and regulations, including the Declaration of Helsinki.


\subsubsection{Segmentation Dataset} 

HR-pQCT scans (XtremeCT II; Scanco Medical AG, Brüttisellen, Switzerland) were acquired at 7.3\% proximal to the tibial endplate using the manufacturer’s standard in vivo protocol (168 slice stack, 60.7 µm nominal isotropic voxel size) \cite{Warden2021}. To improve model robustness and generalizability across acquisition sites, scans were collected from two research centers: the Columbia University Irving Medical Center (CUIMC) and the Musculoskeletal Function, Imaging, and Testing Core (FIT Core) of the Indiana Center for Musculoskeletal Health (ICMH). In total, the segmentation dataset comprised 40 distal tibia HR-pQCT scans from 22 participants (age: 56.12 ± 10.31 years; male: 3, female: 19), yielding a dataset of 6,720 high-resolution 2D images. All scans satisfied image quality criteria, with visual grading scores (VGS) of 1 or 2, indicating minimal motion artifacts \cite{Sode2011}. To prevent data leakage, the dataset was split at the patient level into training (60\%; 4,032 slices from 24 scans), validation (20\%; 1,344 slices from 8 scans), and test (20\%; 1,344 slices from 8 scans) sets. 

\subsubsection{Classification Dataset} The classification dataset comprised HR-pQCT scans from a total of 122 individuals (VGS $\leq$ 2), including 61 patients with confirmed osteoporosis (age: 64.48 ± 5.89, male: 15, female: 46) and 61 age- and sex-matched controls (age: 63.62 ± 6.49, male: 15, female: 46). All scans were acquired at 7.3\% proximal to the tibial endplate using the second-generation XtremeCT II scanner at the FIT Core, ICMH. Osteoporosis status was determined by femoral neck aBMD T-score measured with DXA (Norland Elite; Norland at Swissray, Fort Atkinson, WI), with a T-score $\leq$ –2.5 indicating osteoporosis \cite{Kanis2002}. In total, the dataset included 20,496 HR-pQCT images from 122 scans. Data were split at the patient level into training (80\%) and test (20\%) subsets, yielding 98 scans (16,464 images) for training and 24 scans (4,032 images) for testing. Machine learning models were developed and tuned on the training set using five-fold cross-validation, while the test set was used only once for final performance evaluation on unseen data. 

\subsection{Data Preprocessing}

\subsubsection{Annotation} 

To generate ground truth semantic segmentation maps, the segmentation dataset was annotated using 3D Slicer 5.6.1 (\href{https://www.slicer.org/} {https://www.slicer.org/}), an open-source software for biomedical image segmentation \cite{Fedorov2012}. Each distal tibia scan was manually annotated into five segments: 1) cortical bone of the tibia, 2) trabecular bone of the tibia, 3) cortical bone of the fibula, 4) trabecular bone of the fibula, and 5) soft tissues. For each HR-pQCT scan, manual annotations were performed on every fifth slice using the “Segment Editor” module, and segmentation masks for the intermediate slices were generated through interpolation using the “Fill Between Slices” tool. Annotations were initially performed by a biomedical engineering undergraduate student with approximately 6 months of experience in HR-pQCT image analysis, who was trained and supervised by an expert with over 10 years of experience in musculoskeletal research. The annotations were subsequently reviewed and corrected by a biomedical engineering graduate researcher with 2 years of experience in HR-pQCT imaging, followed by a final expert verification to ensure anatomical consistency and labeling accuracy. The deidentified segmentation dataset, along with the corresponding ground truth masks, will be made publicly available upon publication.

\subsubsection{Standardization}

\begin{figure}[!t]
\centerline{\includegraphics[width=\columnwidth]{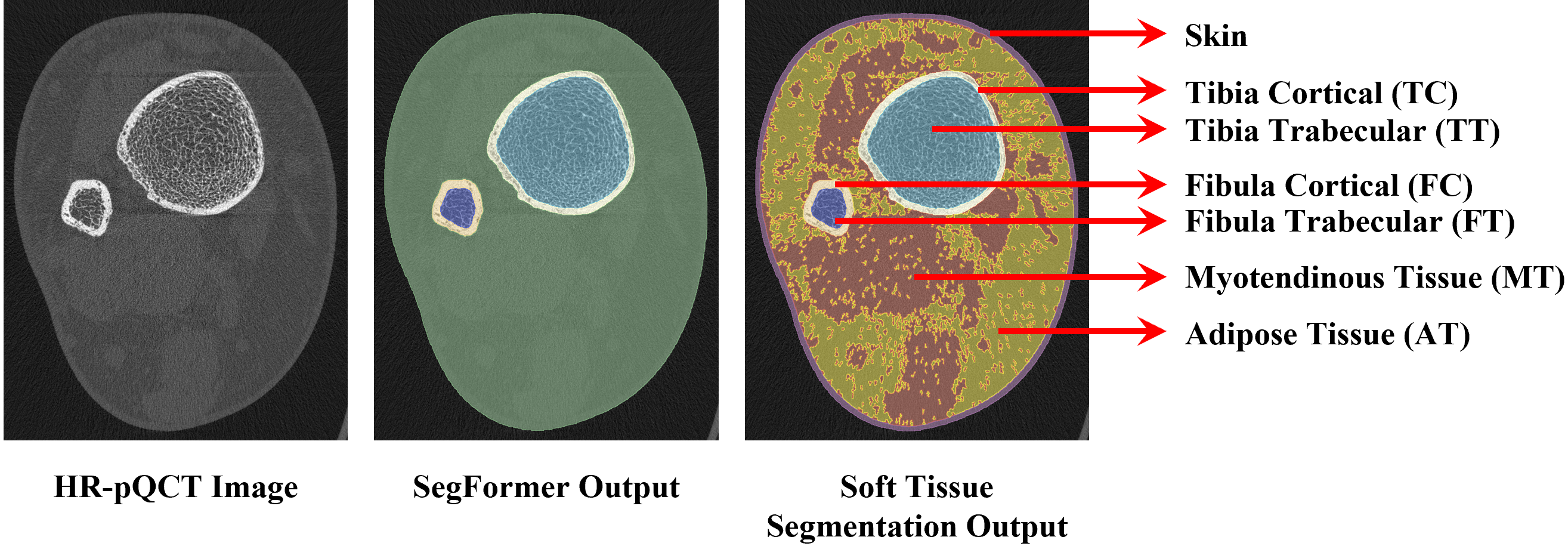}}
\caption{Input HR-pQCT slice to the SegFormer model (left), five region output of the model highlighting the overall soft tissues in green (middle), and refined output with seven distinct regions after soft tissue segmentation (right). The abbreviations indicated in parentheses are used throughout the text.}
\label{fig: segmentation soft tissue}
\end{figure}

The HR-pQCT images and corresponding ground truth masks were cropped to dimensions of 1600 × 1600 pixels to reduce the proportion of background pixels while preserving the anatomical region of interest. Intensity values of the images were clipped to a range of [-4000, 6000] Hounsfield units (HU) to suppress extreme outliers, and the clipped intensities were subsequently normalized to a range of [0, 1] using the transformation in \eqref{eq_x_clip} and \eqref{eq_x_norm}. 

\begin{equation}
x_{clip} = min(6000, max(-4000, x))
\label{eq_x_clip}
\end{equation}

\begin{equation}
x_{norm} = \frac{x_{clip} - min(x_{clip})}{max(x_{clip}) - min(x_{clip})}
\label{eq_x_norm}
\end{equation}

To reduce computational cost, the normalized images were downsampled by a factor of 2 to 800 × 800 pixels using bicubic interpolation, which preserves spatial continuity and fine-grained structural information. The corresponding segmentation masks were resized using nearest neighbor interpolation to maintain label integrity. The resulting dataset was used as input for the deep learning-based segmentation model.  

\subsection{Segmentation}

\subsubsection{Architecture} 

\begin{figure*}[!t]
\centerline{\includegraphics[width=\textwidth]{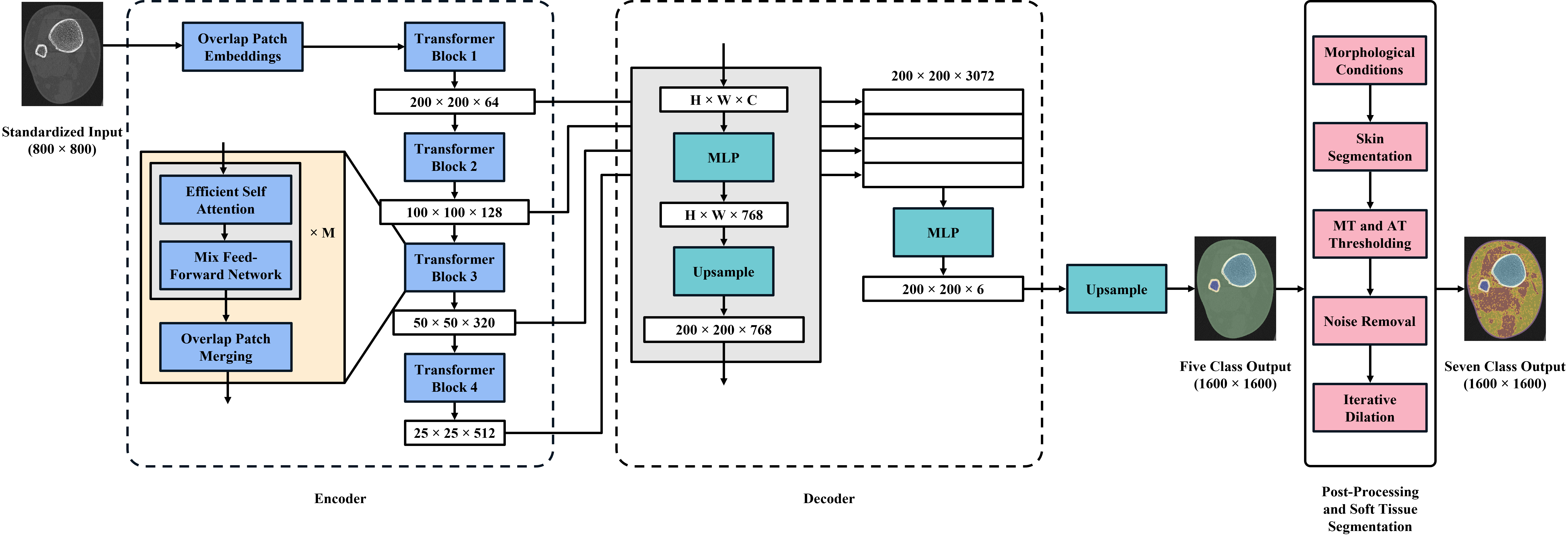}}
\caption{Schematic of the proposed SegFormer-based segmentation pipeline. Standardized HR-pQCT inputs are processed through the encoder-decoder framework to generate a five-class semantic map, followed by post-processing and soft-tissue segmentation to obtain the final seven-class output. For each of the four transformer blocks in the encoder, M denotes the number of repeated efficient self-attention and mix feed-forward network units, with M = 3, 4, 18, and 3. In the decoder, H, W, and C denote the height, width, and channel dimension of the input feature maps.}

\label{fig: Segmentation.png}
\end{figure*}

HR-pQCT images of the distal tibia were segmented into five anatomical regions using the SegFormer model \cite{SegFormer}. SegFormer combines a transformer-based hierarchical encoder with a lightweight multilayer perceptron (MLP) decoder, efficiently capturing both local and global contextual features across multiple spatial scales. 
The SegFormer-B3 variant pretrained on the Cityscapes dataset \cite{Cordts2016} was adapted for grayscale HR-pQCT images by modifying the input patch embedding layer to accept a single channel and initializing its weights by averaging the pretrained RGB channel weights. The model was then fine-tuned on the HR-pQCT dataset, effectively leveraging transfer learning to incorporate prior knowledge from large-scale natural image features. 

Given an input grayscale image of size 800 × 800 × 1, the encoder generates four hierarchical feature representations of sizes 200 × 200 × 64, 100 × 100 × 128, 50 × 50 × 320, and 25 × 25 × 512, respectively (Fig. \ref{fig: Segmentation.png}). These representations capture high-resolution local details in the early stages and low-resolution global semantics in the later stages, enabling the model to learn both fine anatomical boundaries and broader contextual relationships. The decoder projects and upsamples each feature map to a common dimension of 200 × 200 × 768 and concatenates them before further processing through MLP layers. The final output has a size of 200 × 200 × 6, corresponding to the five anatomical regions and the background. During inference, the predicted masks are upsampled to 1600 × 1600 pixels using nearest-neighbor interpolation to match the original HR-pQCT resolution of 60.7 µm. 

All training was performed on an NVIDIA RTX A5500 GPU using Python 3.8.18 and PyTorch 1.11.0. The SegFormer-B3 configuration was chosen for its balance between segmentation accuracy and computational efficiency.

\subsubsection{Training and Optimization} The network was trained for 20 epochs (40,320 iterations) with a batch size of 2 and an initial learning rate of 0.0001 using the Adam optimizer \cite{AdamOptim}. A dynamic learning rate scheduler, ReduceLROnPlateau, was employed to decrease the learning rate by a factor of 10 when the validation Dice score plateaued for 3 consecutive epochs. Gradient clipping with a threshold value of 1 was implemented to prevent exploding gradients during model training. The model was optimized using a combination of equally weighted cross-entropy loss and Dice loss to balance pixel-wise classification accuracy and overlap-based segmentation performance. For a batch of size N, total number of classes C, and total number of pixels per image K, the log-softmax of the model output is computed as

\begin{equation}
\ell_{n,k} = \log \frac{\exp(x_{n,c,k})}{\sum_{i=1}^{C} \exp(x_{n,i,k})},
\label{logsoftmax}
\end{equation}

where $x_{n,c,k}$ denotes the unnormalized logit or raw score assigned by the model to the true class of the n, k-th sample, and $x_{n,i,k}$ is the logit predicted by the model for class i. The cross-entropy loss is then calculated as the negative log-likelihood of the predicted log-probabilities across all pixels and images in a batch.

\begin{equation}
L_{CE} = -\frac{1}{N K} \sum_{n=1}^{N} \sum_{k=1}^{K} \ell_{n,k}
\label{crossentropy}
\end{equation}

To calculate the Dice loss, the raw segmentation outputs were first normalized across the channel dimension using the softmax function. The Dice loss is calculated between the normalized softmax probability of the model $x'_{n,c,k}$ and the one-hot encoded ground truth labels $y'_{n,c,k}$ using \eqref{softmax}, \eqref{dice_score}, and \eqref{dice_loss}. 

\begin{equation}
x'_{n,c,k} = \frac{\exp(x_{n,c,k})}{\sum_{i=1}^{C} \exp(x_{n,i,k})}
\label{softmax}
\end{equation}

\begin{equation}
\ell_{n,c} = \frac{2 \sum_{k=1}^{K} (x'_{n,c,k} \times y'_{n,c,k}) + \varepsilon}
{\sum_{k=1}^{K} x'_{n,c,k} + \sum_{k=1}^{K} y'_{n,c,k} + \varepsilon}
\label{dice_score}
\end{equation}

\begin{equation}
L_{Dice} = 1 - \frac{1}{N C} \sum_{n=1}^{N} \sum_{c=1}^{C} \ell_{n,c}
\label{dice_loss}
\end{equation}

Here, n indexes the images in a batch, c denotes the segmentation class, k indexes the pixels within an image, and $\varepsilon$ is a small constant to prevent division by zero. The Dice score for image n and class c is denoted by $l_{n,c}$, and $L_{Dice}$ is the overall Dice loss across all classes and images. The total segmentation loss is the summation of the cross-entropy and Dice losses.

\begin{equation}
L_{Total} = L_{CE} + L_{Dice}
\label{TotalLoss}
\end{equation}

The final model was selected based on the highest Dice score achieved on the validation dataset after each training epoch. The performance of the segmentation model was evaluated using standard metrics, including precision, recall, F1 score, and intersection over union (IoU), on the test set. For each anatomical region, precision indicates the proportion of positive predictions that are truly positive, while recall measures the probability of correctly detecting a positive class. The F1 score, defined as the harmonic mean of precision and recall, provides a balanced representation of the two. Finally, IoU quantifies the degree of overlap between the predicted segmentation and the ground truth label for each class. For all evaluation metrics, a value of 1 indicates perfect agreement, whereas 0 represents complete disagreement. 

\subsubsection{Post-processing} The segmentation outputs from the deep learning model were refined using a post-processing pipeline that enforced known morphological constraints of the distal tibia site \cite{Neeteson2023}. Briefly, each anatomical class was constrained to a single connected component, with misclassified fragments removed by retaining only the largest instance and reassigning smaller regions based on the majority labels of their neighboring pixels. The continuity of the cortical mask was verified by comparing the area of the cortical contour to the convex hull area, and any discontinuity was corrected using morphological closing operations. Finally, for both the tibia and fibula, the region enclosed within the cortical boundary was reassigned as the corresponding trabecular compartment, ensuring topological accuracy of the final segmentation map. 

\subsubsection{Soft Tissue Segmentation} The soft tissue region of the segmented image was further divided into skin, myotendinous tissue, and adipose tissue using a modified soft tissue analysis protocol for second-generation HR-pQCT scans, implemented in Python 3.8.18 \cite{Erlandson2017, Hildebrand2021}. Initially, a 2 mm-wide band along the outer boundary of the soft tissue mask was designated as the skin and excluded from subsequent analysis. Within the remaining soft tissue mask, seed regions were planted based on manufacturer-recommended thresholds of 100 to 600 HU for myotendinous tissue and –600 to –200 HU for adipose tissue. Seeds smaller than 30 pixels were removed from the image to reduce noise, and the remaining seeds were iteratively dilated by one pixel for 20 iterations. The pixels that remained unassigned after region growing and the pixels with overlapping seeds were subsequently resolved using a –50 HU threshold, where pixels above this value were classified as myotendinous tissue and those below as adipose tissue. The final segmentation mask thus delineates myotendinous tissue, adipose tissue, and skin, in addition to the bone compartments, as shown in Fig. \ref{fig: segmentation soft tissue} and Fig. \ref{fig: Segmentation.png}. 

\subsection{Radiomics Feature Analysis} 
\subsubsection{Feature Extraction} Following segmentation, radiomic features were extracted from 2D images using PyRadiomics 3.0.1 \cite{vanGriethuysen2017}. Features were calculated from both the original HR-pQCT images and filtered versions of the original images. Eight types of image filters were applied to capture diverse intensity and texture characteristics: (1) Laplacian of Gaussian (LoG) filter ($\sigma = 2$); (2) wavelet filter with one level of decomposition; (3) square filter; (4) square root filter; (5) logarithm filter; (6) exponential filter; (7) gradient filter; and (8) local binary pattern (LBP) filter with eight neighbors.  

Seven classes of radiomic features were calculated, including 18 first-order statistical features, 9 2D shape features, 24 gray level co-occurrence matrix (GLCM) features, 16 gray level size zone matrix (GLSZM) features, 16 gray level run length matrix (GLRLM) features, 5 neighboring gray tone difference matrix (NGTDM) features, and 14 gray level dependence matrix (GLDM) features. The 9 shape features were derived solely from the segmentation masks and are independent of image intensity. The remaining 93 radiomic features were extracted from the original and filtered image sets, totaling 939 features for a single anatomical region, as shown in Fig. \ref{fig: Feature Extraction and Classification}(a). 

\subsubsection{Feature Selection} Radiomic features extracted from 20,496 images were divided on a patient level into training (16,464 images) and testing (4,032 images) sets for subsequent machine learning prediction. All features were normalized to a range of [0, 1], and dimensionality reduction techniques were performed to minimize data redundancy, mitigate potential overfitting, and improve model generalizability. Three complementary feature selection methods were used to achieve robust and unbiased feature reduction prior to classification: variance thresholding (threshold = 0.02), correlation analysis (Pearson’s $|r| > 0.9$), and least absolute shrinkage and selection operator (LASSO) regression (Fig. \ref{fig: Feature Extraction and Classification}(a)). These steps substantially reduced feature dimensionality by eliminating features with minimal variance across the samples, removing highly correlated variables, and retaining only the most informative predictors. Feature selection was performed on the training set, and the same transformations were applied to the test set.

\subsection{Image Level Osteoporosis Prediction} 

\begin{figure*}[!t]
\centerline{\includegraphics[width=\textwidth]{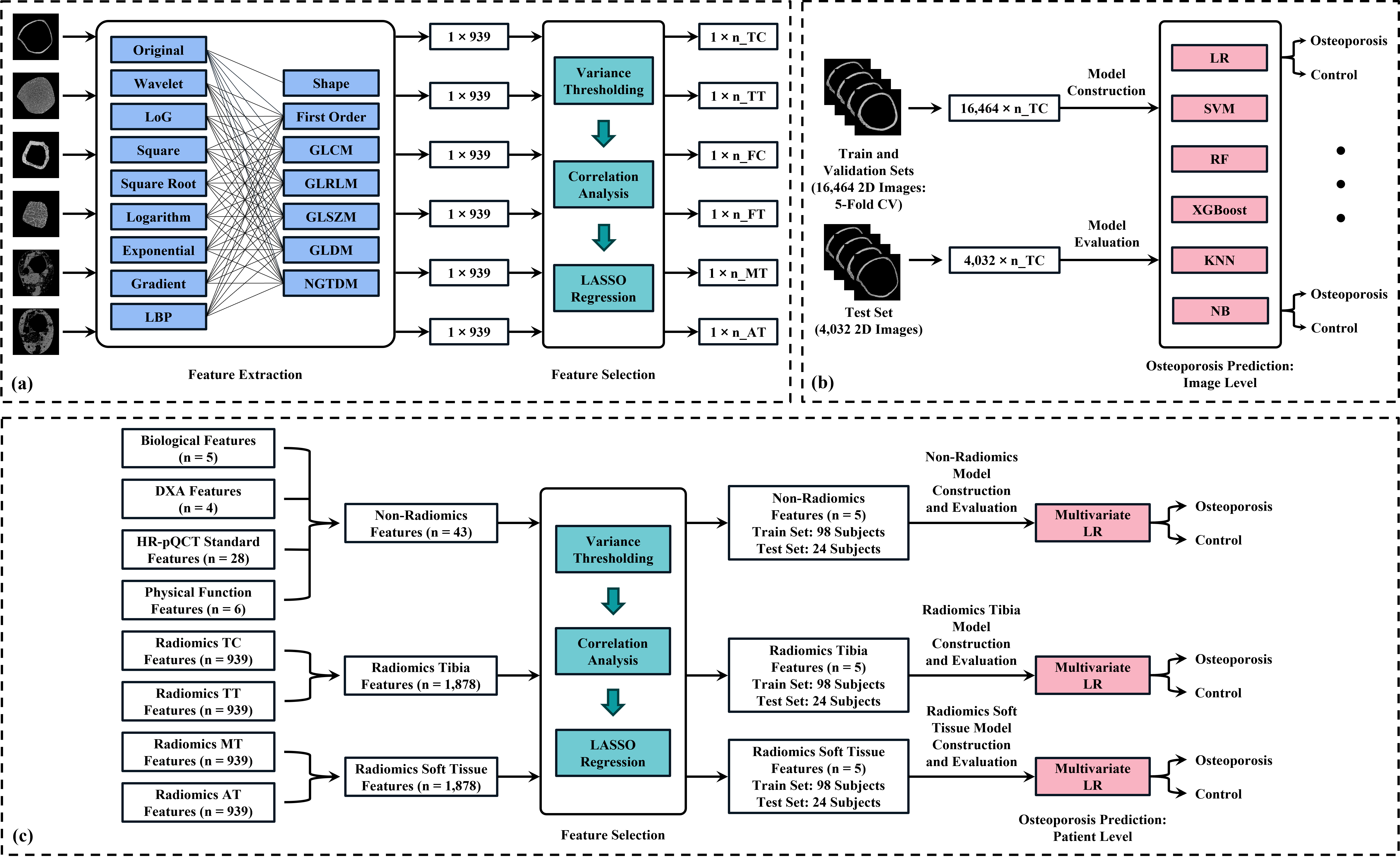}}
\caption{Overview of the proposed osteoporosis classification framework. (a) From each of the seven segmented regions except skin, 939 radiomics features are extracted and reduced through feature selection. (b) Region-specific models are trained using six machine learning classifiers with five-fold cross-validation on 16,464 images and tested on 4,032 held-out images. (c) Different features are grouped into three subsets: non-radiomics, radiomics tibia, and radiomics soft-tissue. Selected features from the subsets are used to construct patient level models (98 training, 24 testing subjects) for binary osteoporosis prediction.}

\label{fig: Feature Extraction and Classification}
\end{figure*}

After feature selection, the remaining most informative radiomics features were used to train six machine learning classifiers, including linear models (logistic regression), kernel-based approaches (support vector machine), distance-based methods (K-nearest neighbors), probabilistic models (naive Bayes), and tree-based ensemble frameworks (random forest,  extreme gradient boosting) (Fig. \ref{fig: Feature Extraction and Classification}(b)). Hyperparameter tuning for each classifier was conducted through a grid search procedure, in which candidate parameter combinations were evaluated using 5-fold cross-validation on the training set comprising 16,464 HR-pQCT images from 98 subjects. This approach ensured that model optimization relied solely on internal validation within the training data, while the final predictive performance was assessed on an independent test set of 4,032 images from 24 subjects, thereby preventing data leakage and providing an unbiased estimate of generalization. 

The classification performance was quantified using accuracy, sensitivity, specificity, F1 score, and the area under the receiver operating characteristic curve (AUROC). Accuracy measures the overall proportion of correctly classified images. Sensitivity represents the probability of correctly identifying osteoporosis cases, while specificity quantifies the model’s ability to correctly identify controls. The ROC curve depicts the trade-off between the true positive rate (TPR) and the false positive rate (FPR) across classification thresholds, with AUROC summarizing this relationship. A higher AUROC value (approaching 1) indicates strong discriminative performance, while a value near 0.5 reflects random classification.

\subsection{Patient Level Osteoporosis Prediction} 
\subsubsection{Feature Extraction} To evaluate the predictive value of radiomics features relative to established biological, functional, and clinical measures, multiple categories of patient-level features were extracted, as shown in Fig. \ref{fig: Feature Extraction and Classification}(c). Biological parameters included age, sex, body mass index (BMI) (kg/m$^2$), tibia length (mm), and ulna length (mm) \cite{Warden2025}. Functional outcomes comprised both self-reported and performance-based assessments, including the physical function domain of the National Institutes of Health patient reported outcomes measurement information system (PROMIS-PF) (performed via computerized adaptive testing) \cite{Cella2007}, usual gait speed (m/s), fast gait speed (m/s), grip strength (kg), five-time repeated sit-to-stand (R-STS) chair test (s), and six-minute walk distance (m), capturing overall strength, mobility, and coordination \cite{Warden2025}. DXA-derived features were obtained from whole-body and regional scans, excluding BMD and T-score variables used for osteoporosis diagnosis. Specifically, the retained DXA features were whole-body total and subtotal percent fat mass, appendicular lean mass (g), and appendicular lean mass/height$^2$ (kg/m$^2$). Standard HR-pQCT parameters assessed bone microarchitecture and density, including cortical and trabecular thickness (mm), cortical porosity (\%), trabecular number (1/mm), and cortical and trabecular volumetric BMD (mgHA/cm$^3$). Micro finite element analysis (\textmu FEA) estimated the mechanical properties of bone, such as stiffness (N/mm) and ultimate failure load (N) \cite{Warden2021}. In total, 28 standard HR-pQCT bone features were included in the present study. 

\begin{figure*}[!t]
\centerline{\includegraphics[width=\textwidth]{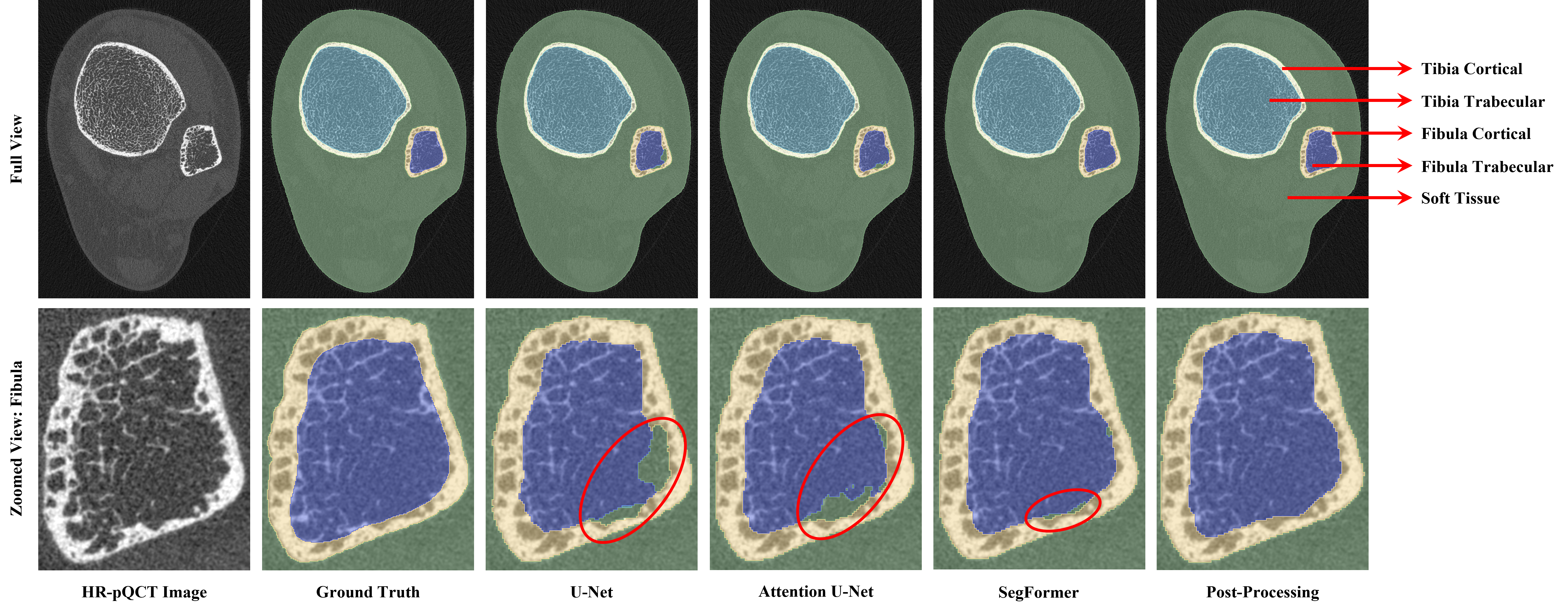}}
\caption{Qualitative semantic segmentation performance on a sample HR-pQCT image (7.3\% distal tibia region). The top row shows the full HR-pQCT image encompassing all five regions, and the bottom row highlights the fibula. U-Net–based architectures demonstrate noticeable pixel misclassification, which is substantially reduced with the SegFormer model (highlighted in red). The remaining minor inconsistencies are addressed through post-processing.}
\label{fig: segmentation model comparison}
\end{figure*}

Together, the biological, functional, DXA, and HR-pQCT variables formed the non-radiomics feature group, comprising 43 features in total. To ensure comparability with this group, radiomics features were averaged across all 168 images for each participant, resulting in patient-level representations. Consistent with the image-level workflow, each subject had 939 radiomics features calculated separately for the cortical and trabecular bone of the tibia and fibula, as well as for the myotendinous and adipose tissue regions. 

\subsubsection{Feature Selection and Statistical Analysis}
The dataset was divided into training (98 subjects) and testing (24 subjects) sets, consistent with the image-level analysis. Feature selection was conducted on the normalized training set following a similar three-stage pipeline: variance thresholding, correlation analysis, and LASSO regression (Fig. \ref{fig: Feature Extraction and Classification}(c)). Given the limited number of training subjects, only the top five features based on absolute LASSO coefficients were selected to further mitigate overfitting. The resulting subset of features was used to construct several multivariate logistic regression models: (1) a baseline model incorporating non-radiomics variables, (2) a radiomics tibia model including features from tibial cortical and trabecular compartments, and (3) a radiomics soft tissue model including features from myotendinous and adipose tissues. Model calibration was evaluated using the Hosmer-Lemeshow goodness-of-fit test. The test remains a standard benchmark for logistic regression calibration in radiomics literature \cite{Wang2023}, and was interpreted alongside classification performance metrics to provide a holistic view of model reliability given the modest sample size. Predictive performance was assessed on the held-out test set using accuracy, sensitivity, specificity, F1 score, and AUROC. The optimal decision threshold for classifying osteoporosis was determined using the Youden index. All statistical analyses were performed in R (version 4.4.1). 

\section{Results}
\subsection{Segmentation Results}

To evaluate the performance of the proposed SegFormer framework against conventional CNNs, comparisons were made with U-Net \cite{Ronneberger2015} and Attention U-Net \cite{oktay2018attention} models. Quantitative results for all models are presented in Table \ref{tab: segmentation model comparison}, which summarizes the mean and standard deviation (SD) of the segmentation metrics (precision, recall, F1 score, and IoU) computed across all test samples of the segmentation dataset. 

\begin{table}[ht!]
\centering
\caption{Segmentation performance of different models across anatomical structures (mean $\pm$ SD). Bold values indicate the best performance per structure.}
\renewcommand{\arraystretch}{1.2}
\setlength{\tabcolsep}{2pt}
\begin{tabular}{|llcccc|}
\hline
\textbf{Model} & \makecell{\textbf{Anatomical}\\\textbf{Structure}} & \textbf{Precision} & \textbf{Recall} & \textbf{F1 Score} & \textbf{IoU} \\ 
\hline
\multirow{5}{*}{U-Net} 
& Soft Tissues & 98.4$\pm$7.3 & 99.6$\pm$1.7 & 98.8$\pm$5.5 & 98.0$\pm$7.5 \\
& Tibia Cortical & 93.2$\pm$5.3 & \textbf{91.7$\pm$3.6} & 92.4$\pm$3.9 & 86.1$\pm$6.4 \\
& Tibia Trabecular & 98.8$\pm$1.0 & 99.4$\pm$2.0 & 99.1$\pm$1.6 & 98.2$\pm$2.1 \\
& Fibula Cortical & 83.8$\pm$16.3 & 89.2$\pm$11.9 & 85.5$\pm$12.9 & 76.7$\pm$18.1 \\
& Fibula Trabecular & \textbf{97.0$\pm$4.0} & 77.2$\pm$25.5 & 82.9$\pm$18.2 & 74.4$\pm$23.4 \\ 
\hline
\multirow{5}{*}{\makecell{Attention\\U-Net}} 
& Soft Tissues & 98.6$\pm$5.8 & 99.6$\pm$2.3 & 98.9$\pm$4.6 & 98.2$\pm$6.2 \\
& Tibia Cortical & \textbf{94.8$\pm$3.3} & 91.4$\pm$3.6 & \textbf{92.9$\pm$2.7} & \textbf{87.0$\pm$4.6} \\
& Tibia Trabecular & \textbf{98.9$\pm$0.6} & 99.4$\pm$1.3 & \textbf{99.2$\pm$0.8} & \textbf{98.3$\pm$1.3} \\
& Fibula Cortical & 86.4$\pm$14.0 & 91.5$\pm$5.5 & 88.2$\pm$8.3 & 79.8$\pm$12.5 \\
& Fibula Trabecular & 95.3$\pm$4.8 & 76.5$\pm$27.9 & 80.9$\pm$20.2 & 72.1$\pm$24.7 \\ 
\hline
\multirow{5}{*}{SegFormer} 
& Soft Tissues & \textbf{99.6$\pm$0.1} & \textbf{99.6$\pm$0.1} & \textbf{99.6$\pm$0.1} & \textbf{99.2$\pm$0.2} \\
& Tibia Cortical & 94.7$\pm$2.8 & 90.9$\pm$4.0 & 92.7$\pm$2.1 & 86.5$\pm$3.6 \\
& Tibia Trabecular & 98.8$\pm$0.8 & \textbf{99.4$\pm$0.3} & 99.1$\pm$0.3 & 98.2$\pm$0.6 \\
& Fibula Cortical & \textbf{89.6$\pm$11.0} & \textbf{93.5$\pm$3.3} & \textbf{91.0$\pm$5.5} & \textbf{83.9$\pm$8.7} \\
& Fibula Trabecular & 96.0$\pm$4.1 & \textbf{93.5$\pm$7.4} & \textbf{94.4$\pm$3.3} & \textbf{89.6$\pm$5.6} \\ 
\hline
\end{tabular}
\label{tab: segmentation model comparison}
\end{table}

\begin{figure*}[!t]
\centerline{\includegraphics[width=\textwidth]{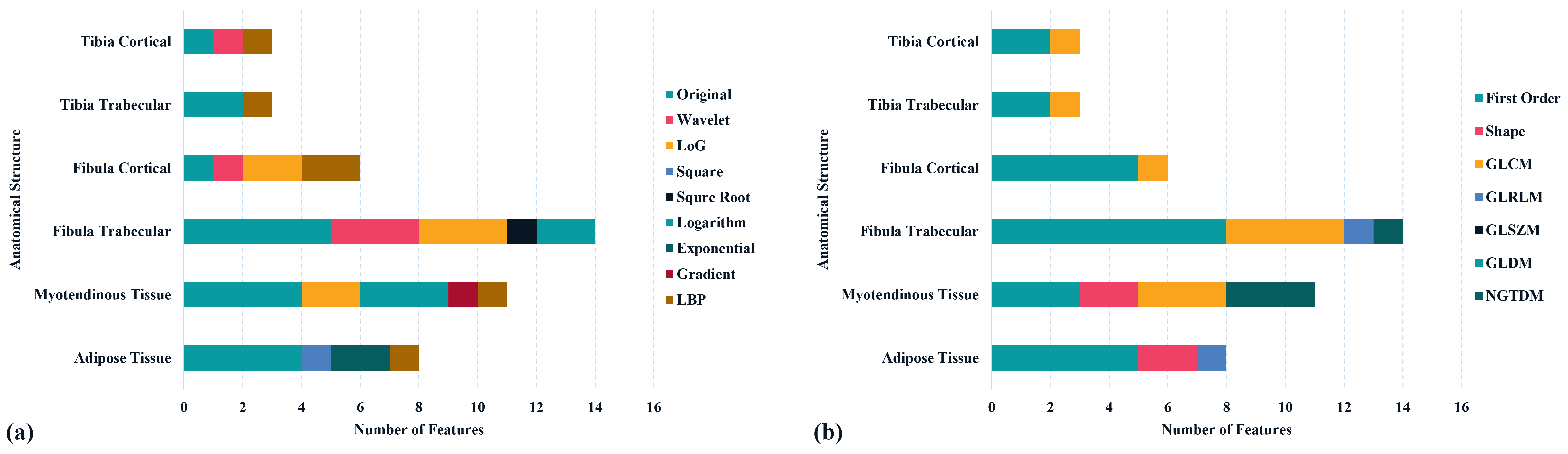}}
\caption{Retained features after the feature selection process. (a) Features grouped by filter classes. (b) Features grouped by feature classes.}
\label{fig: selected features image}
\end{figure*}

For both U-Net and Attention U-Net, segmentation performance was lowest in the fibula trabecular region, with IoU values of $74.4\pm23.4$ and $72.1\pm24.7$, respectively. These models exhibited high precision but low recall with substantial inter-image variability, suggesting a tendency to produce false negatives and under-segment the target structure. Similar performance degradation was observed for the fibula cortical region, indicating limited generalization to minority classes with fewer pixels. In contrast, both models achieved near-perfect segmentation for larger classes such as soft tissue and tibia trabecular bone, with Attention U-Net reaching mean IoU scores of $98.2\pm6.2$ and $98.3\pm1.3$ on the unseen test set, respectively. 

The SegFormer model markedly improved segmentation accuracy in the minority classes, yielding mean IoU gains of 5.14\% and 20.43\% for the fibula cortical and trabecular regions, respectively, compared to the second-best model. Across all regions, it demonstrated the lowest inter-image variability (3.74\% in IoU), outperforming U-Net (11.5\%) and Attention U-Net (9.86\%). SegFormer also achieved the highest performance in the soft tissue region and ranked marginally behind Attention U-Net in tibia segmentation. These findings highlight the effectiveness of transformer-based architectures in modeling global contextual relationships across anatomical boundaries while preserving fine local structural details. 

Figure \ref{fig: segmentation model comparison} illustrates representative qualitative segmentation results on the test set. U-Net–based models demonstrated visible under-segmentation and trabecular misclassification due to intensity similarities between deteriorated trabeculae and adjacent soft tissues (Fig. \ref{fig: segmentation model comparison}, bottom row). In contrast, SegFormer provided substantially improved segmentation, with only minor pixel-level inconsistencies that were further refined through post-processing. The final segmentation output was subsequently processed using a modified soft tissue analysis protocol as described in Section III, subdividing the soft tissue into skin, myotendinous tissue, and adipose tissue (Fig. \ref{fig: segmentation soft tissue}). 

\subsection{Radiomics Feature Analysis}

From each segmented region except skin, 939 radiomics features were extracted and refined through a feature-selection pipeline, as described in Section III. Variance thresholding reduced the feature count by approximately half (356 to 623 features), while correlation analysis yielded a tenfold reduction (36 to 59 features). Following LASSO regression, the final number of retained features varied between 3 and 14, depending on the region.

Fig. \ref{fig: selected features image} summarizes the selected features by filter and feature types. As presented in Fig. \ref{fig: selected features image}(a), most selected features were calculated from the original HR-pQCT images, followed by those derived from wavelet, LoG, logarithm, and LBP filters. The wavelet filter performed single-level decomposition to capture both low-frequency approximations and high-frequency detail components. The LoG filter enhanced edge information by emphasizing areas with rapidly changing intensities. The logarithm filter improved soft tissue contrast within the image, whereas the LBP filter encoded local texture information. As shown in Fig. \ref{fig: selected features image}(b), among the 45 retained features across all six regions, 25 were first-order features, followed by 10 GLCM features. First-order features characterize pixel intensity distributions using statistical descriptors such as energy, entropy, skewness, and kurtosis. GLCM features, derived from the joint probability distribution of pixel pairs separated by a defined spatial offset \cite{Haralick1973}, quantify textural attributes including coarseness, local intensity variation, and gray-level inhomogeneity. Collectively, these features capture multi-scale intensity, texture, spatial orientation, and frequency characteristics of the HR-pQCT images. 

\begin{figure}[h]
\centerline{\includegraphics[width=\columnwidth]{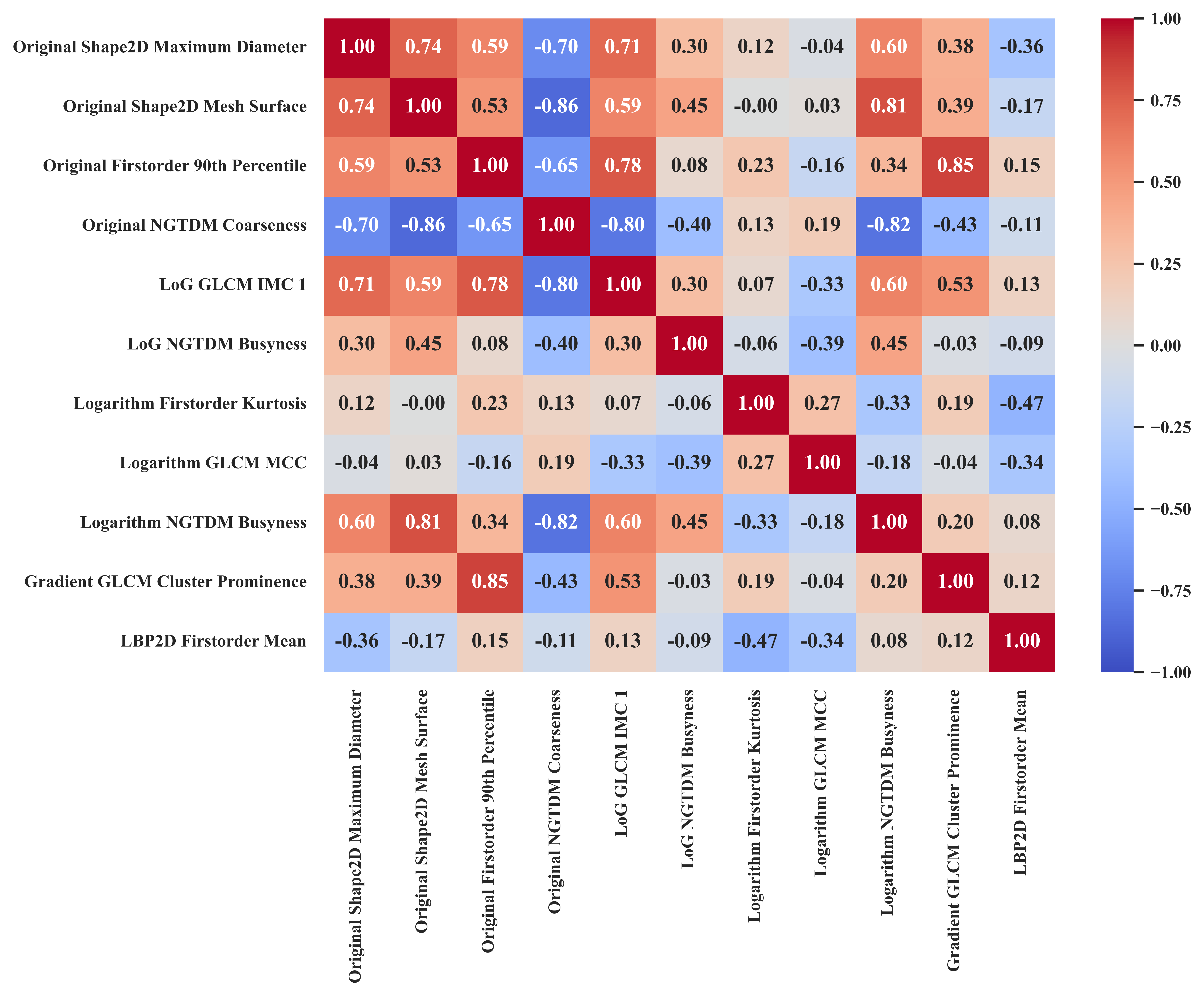}}
\caption{Correlation heatmap of features extracted from the myotendinous tissue region. The mean absolute correlation coefficient of the selected features was 0.358.}
\label{fig: correlation heatmap}
\end{figure}

The correlation heatmap of the selected features for the myotendinous tissue region shows a mean absolute correlation coefficient of 0.358 (Fig. \ref{fig: correlation heatmap}). Across all six regions, mean absolute correlation values ranged from 0.358 to 0.526, indicating moderate independence among the retained features. 

\subsection{Image Level Prediction Results} 
    
\begin{table}[b!]
\centering
\begin{threeparttable}
\caption{Classification performance of different models across anatomical structures on the held-out test set. Bold values indicate the best performance per structure.}
\renewcommand{\arraystretch}{1.2}
\setlength{\tabcolsep}{4pt}
\begin{tabular}{|llccccc|}
\hline
\makecell{\textbf{Anatomical}\\\textbf{Structure}} & \textbf{Model\tnote{a}} & \textbf{Acc.} & \textbf{Sen.} & \textbf{Spe.} & \textbf{F1 Score} & \textbf{AUROC} \\ 

\hline
\multirow{6}{*}{\makecell{Tibia\\Cortical}} 

& LR & 76.69 & 64.34 & 89.04 & 0.734 & 0.777 \\
& SVM & 73.59 & 57.44 & \textbf{89.73} & 0.685 & 0.766 \\
& RF & 68.53 & \textbf{69.05} & 68.01 & 0.687 & 0.777 \\
& XGBoost & 73.64 & 62.45 & 84.82 & 0.703 & 0.777 \\
& KNN & 69.25 &	50.35 & 88.14 & 0.621 & 0.769 \\
& NB & \textbf{77.48} & 67.36 & 87.60 & \textbf{0.749} & \textbf{0.792} \\ 

\hline
\multirow{6}{*}{\makecell{Tibia\\Trabecular}} 
& LR & \textbf{78.55} & 67.51 & 89.58 & \textbf{0.759} & \textbf{0.799} \\
& SVM & 75.57 & 61.36 & 89.78 & 0.715 & 0.788 \\
& RF & 72.59 & \textbf{73.61} & 71.58 & 0.729 & 0.777 \\
& XGBoost & 70.51 & 72.67 & 68.35 & 0.711 & 0.774 \\
& KNN & 73.14 & 73.07 & 73.21 & 0.731 & 0.792 \\
& NB & 75.57 & 59.38 & \textbf{91.77} & 0.709 & 0.794 \\ 

\hline
\multirow{6}{*}{\makecell{Fibula\\Cortical}} 
& LR & \textbf{78.99} & 67.26 & 90.72 & \textbf{0.762} & \textbf{0.847} \\
& SVM & 78.27 & 61.21 & 95.34 & 0.738 & 0.842 \\
& RF & 69.84 & 50.84 & 88.84 & 0.628 & 0.733 \\
& XGBoost & 74.85 & \textbf{68.85} & 80.85 & 0.733 & 0.791 \\
& KNN & 68.75 & 64.68 & 72.82 & 0.674 & 0.739 \\ 
& NB & 78.94 & 62.25 & \textbf{95.63} & 0.747 & 0.839 \\ 

\hline
\multirow{6}{*}{\makecell{Fibula\\Trabecular}} 
& LR & 64.48 & 44.10 & 84.87 & 0.554 & 0.679 \\
& SVM & 65.13 & 45.73 & 84.52 & 0.567 & 0.676 \\
& RF & \textbf{67.93} & 38.34 & \textbf{97.52} & 0.545 & 0.705 \\
& XGBoost & 66.64 & 40.53 & 92.76 & 0.549 & \textbf{0.719} \\
& KNN & 63.94 & 38.34 & 89.53 & 0.515 & 0.694 \\ 
& NB & 67.86 & \textbf{52.83} & 82.89 & \textbf{0.622} & 0.709 \\ 

\hline
\multirow{6}{*}{\makecell{Myotendinous\\Tissue}} 
& LR & \textbf{80.08} & \textbf{73.71} & 86.46 & \textbf{0.787} & \textbf{0.850} \\
& SVM & 79.44 & 72.17 & \textbf{86.71} & 0.778 & 0.848 \\
& RF & 74.68 & 66.27 & 83.09 & 0.724 & 0.785 \\
& XGBoost & 75.64 & 68.25 & 83.04 & 0.737 & 0.848 \\
& KNN & 73.02 & 69.30 & 76.74 & 0.720 & 0.758 \\ 
& NB & 75.07 & 66.62 & 83.53 & 0.728 & 0.780 \\ 

\hline
\multirow{6}{*}{\makecell{Adipose\\Tissue}} 
& LR & 77.73 & 70.49 & 84.97 & 0.760 & \textbf{0.857} \\
& SVM & \textbf{78.50} & 71.73 & 85.27 & \textbf{0.769} & 0.833 \\
& RF & 70.86 & 67.21 & 74.50 & 0.698 & 0.710 \\
& XGBoost & 74.98 & 58.63 & \textbf{91.32} & 0.701 & 0.808 \\
& KNN & 67.11 & \textbf{77.28} & 56.94 & 0.702 & 0.658 \\ 
& NB & 73.44 & 56.00 & 90.87 & 0.678 & 0.721 \\ 

\hline
\end{tabular}

\begin{tablenotes}
\footnotesize
\item[a] LR: logistic regression; SVM: support vector machine; RF: random forest; XGBoost: extreme gradient boosting; KNN: k-nearest neighbors; NB: naive Bayes.
\end{tablenotes}

\label{tab: classification model comparison region (image)}
\end{threeparttable}
\end{table}

The selected radiomics features were employed to classify osteoporosis and control groups across six anatomical regions using six machine learning classifiers. As summarized in Table \ref{tab: classification model comparison region (image)}, logistic regression consistently demonstrated superior or near-superior performance in terms of classification accuracy, F1 score, and AUROC, on the held-out test set across most regions. This outcome aligns with the LASSO-based feature selection process, which inherently favors linearly separable features, thereby enhancing logistic regression performance. The remaining classifiers exhibited region-dependent variability in predictive accuracy.

\begin{figure}[h]
\centerline{\includegraphics[width=\columnwidth]{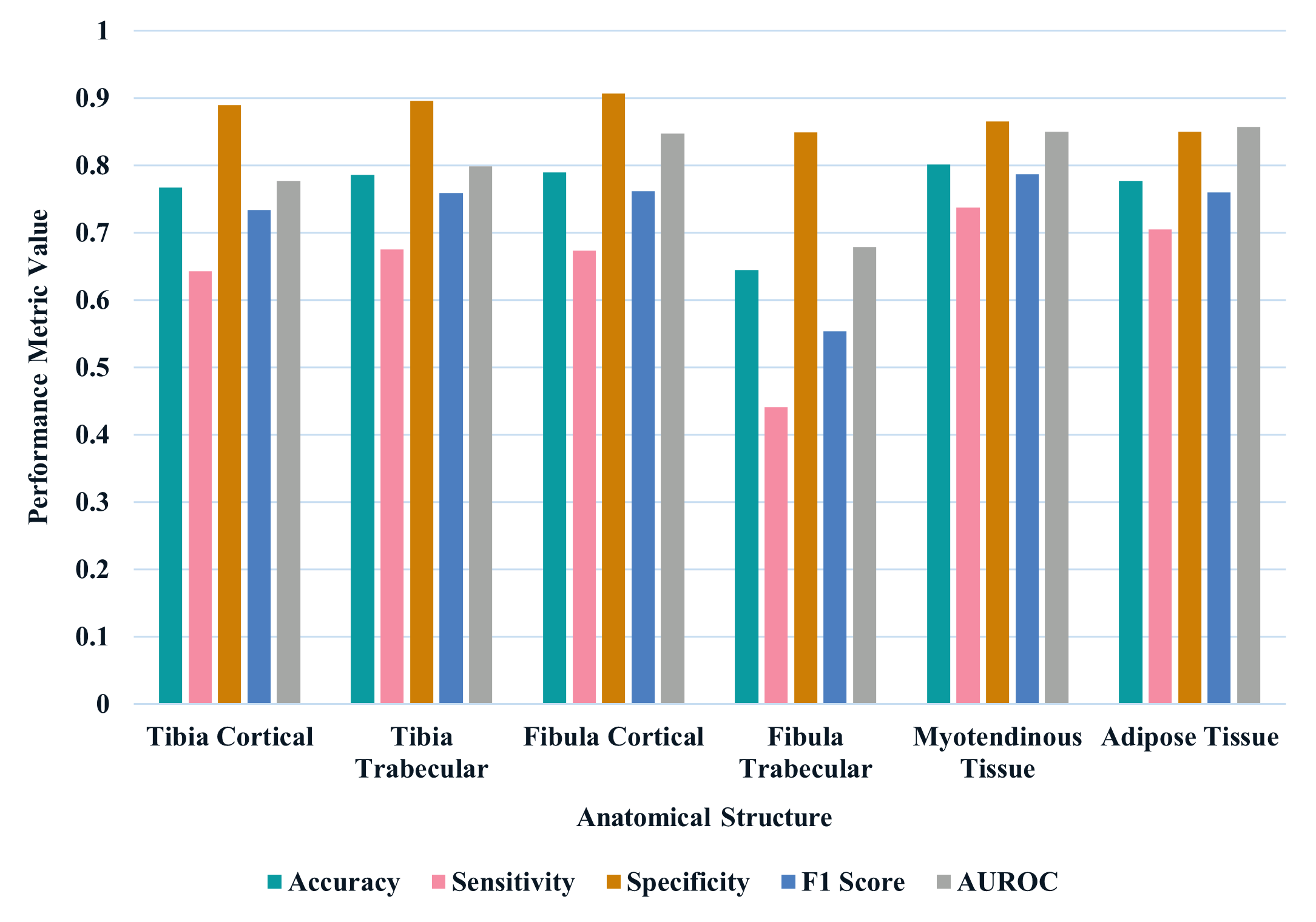}}
\caption{Classification performance of the Logistic Regression model using features extracted from different anatomical regions.}
\label{fig: classification bar graph - LR}
\end{figure}

Fig. \ref{fig: classification bar graph - LR} illustrates the regional performance of the logistic regression model. The highest accuracy (80.08\%) and F1 score (0.787) were achieved using myotendinous tissue features, while the best AUROC (0.857) was obtained from adipose tissue, closely followed by myotendinous tissue (0.850). These results indicate that soft-tissue regions contain strong discriminative information for distinguishing subject groups beyond the traditionally analyzed bone compartments. Among bone compartments, the tibia trabecular region outperformed the tibia cortical region, whereas features extracted from the fibula trabecular region showed limited predictive performance. This potentially reflects slower trabecular deterioration within the fibula and greater overlap between osteoporotic and control patterns. For all regions, logistic regression favored specificity over sensitivity, resulting in more false negatives. However, this balance was improved in soft-tissue regions, suggesting greater separability of control and osteoporotic samples in these tissues. 

\begin{table}[h]
\centering
\caption{Classification performance of different models across soft tissue regions at varying radial distances from the tibia on the held-out test set. Bold values indicate the best performance per region.}
\renewcommand{\arraystretch}{1.2}
\setlength{\tabcolsep}{4pt}
\begin{tabular}{|llccccc|}
\hline
\makecell{\textbf{Anatomical}\\\textbf{Structure}} & \textbf{Model} & \textbf{Acc.} & \textbf{Sen.} & \textbf{Spe.} & \textbf{F1 Score} & \textbf{AUROC} \\ 

\hline
\multirow{6}{*}{\makecell{Soft Tissue\\(10 mm)}} 

& LR & 77.60 & 60.96 & 94.25 & 0.731 & 0.844 \\
& SVM & 77.78 & 61.21 & \textbf{94.35} & 0.734 & 0.835 \\
& RF & 77.65 & \textbf{87.65} & 67.66 & \textbf{0.797} & 0.850 \\
& XGBoost & \textbf{78.17} & 83.58 & 72.77 & 0.793 & \textbf{0.875} \\
& KNN & 71.65 & 59.77 & 83.53 & 0.678 & 0.793 \\ 
& NB & 74.36 & 59.57 & 89.14 & 0.699 & 0.809 \\ 

\hline
\multirow{6}{*}{\makecell{Soft Tissue\\(20 mm)}} 
& LR & \textbf{77.55} & 78.42 & 76.69 & \textbf{0.778} & \textbf{0.792} \\
& SVM & 76.96 & \textbf{79.46} & 74.45 & 0.775 & 0.790 \\
& RF & 70.16 & 71.83 & 68.50 & 0.707 & 0.774 \\
& XGBoost & 71.45 & 62.10 & \textbf{80.80} & 0.685 & 0.757 \\
& KNN & 72.10 & 63.64 & 80.56 & 0.695 & 0.784 \\ 
& NB & 74.45 & 79.32 & 69.59 & 0.756 & 0.772 \\ 

\hline
\multirow{6}{*}{\makecell{Soft Tissue\\(Full)}} 
& LR & \textbf{77.78} & 65.53 & \textbf{90.03} & 0.747 & \textbf{0.822} \\
& SVM & 77.08 & 72.82 & 81.35 & \textbf{0.761} & 0.815 \\
& RF & 72.92 & 69.05 & 76.79 & 0.718 & 0.740 \\
& XGBoost & 73.07 & 74.21 & 71.92 & 0.734 & 0.815 \\
& KNN & 71.11 & \textbf{91.32} & 50.89 & 0.760 & 0.750 \\ 
& NB & 74.11 & 81.85 & 66.37 & 0.760 & 0.766 \\ 

\hline
\end{tabular}
\label{tab: classification model comparison radial (image)}
\end{table}

In addition to the segmented regions, the impact of soft tissues as a function of distance from the tibia was analyzed. For this purpose, radiomics features were extracted from concentric soft tissue regions defined by radial distances from the outer surface of the tibia (10 mm and 20 mm), as well as from the entire soft-tissue compartment (myotendinous tissue, adipose tissue, and skin combined). The selected features were then used to classify osteoporosis using multiple machine learning models, and the corresponding results are summarized in Table \ref{tab: classification model comparison radial (image)}. Among all models, XGBoost demonstrated the best performance within the 10 mm region, achieving an accuracy of 78.17\%, an F1 score of 0.793, and an AUROC of 0.875. This suggests that soft tissues located closer to the bone contain stronger and more localized discriminatory signals associated with osteoporosis status. Logistic regression achieved AUROC values of 0.792 and 0.822 for the 20 mm and full soft-tissue regions, respectively, further supporting the predictive relevance of soft-tissue-derived radiomics features. 

\subsection{Patient Level Prediction Results}

To compare the performance of radiomics features with conventional measures in classifying osteoporosis, three multivariate logistic regression models were developed at the patient level: a non-radiomics model, a radiomics tibia model, and a radiomics soft-tissue model. For each model, the top five features with the highest absolute LASSO coefficients were retained. Model calibration assessed via the Hosmer-Lemeshow test yielded p-values of 0.248, 0.399, and 0.719, respectively, indicating good fit in the training set. Table \ref{tab: classification model comparison stat (patient)} summarizes the results of the multivariate logistic regression models. Across the three models, the variance inflation factor (VIF) varied from 1.135 to 2.041, suggesting minimal multicollinearity among the retained variables. 

\begin{table}[h]
\centering
\begin{threeparttable}
\caption{Results of the logistic regression models, including regression slope coefficients ($\beta$), odds ratios (OR), 95$\%$ confidence intervals (CI), variance inflation factors (VIF), and corresponding p-values for the selected variables.}

\renewcommand{\arraystretch}{1.4}
\setlength{\tabcolsep}{2pt}
\begin{tabular}{|l|l|c|c|c|c|}
\hline
\textbf{Model} & \textbf{Feature Name\tnote{a}} & $\boldsymbol{\beta}$ & \textbf{OR (95$\%$ CI)} & \textbf{\textit{p} Value} & \textbf{VIF} \\ 

\hline
\multirow{5}{*}{\makecell{Non-\\Radiomics}}

& \makecell[l]{Whole Body Total \\ \% Fat Mass} & -0.744 & \makecell[c]{0.475 \\ (0.257, 0.822)} & 0.011\textsuperscript{\textbf{*}} & 1.236 \\
\cline{2-6}

& TbN & -0.761 & \makecell[c]{0.467 \\ (0.229, 0.873)} & 0.024\textsuperscript{\textbf{*}} & 1.135 \\
\cline{2-6}

& CtvBMD & -1.082 & \makecell[c]{0.339 \\ (0.151, 0.684)} & 0.004\textsuperscript{\textbf{*}} & 1.872 \\
\cline{2-6}

& Stiffness & -0.895 & \makecell[c]{0.409 \\ 
(0.176, 0.868)} & 0.026\textsuperscript{\textbf{*}} & 1.393 \\
\cline{2-6}

& (Tb.F/TF)prox & -0.670 & \makecell[c]{0.512 \\ (0.242, 1.000)} & 0.061 & 1.881 \\ 

\hline
\multirow{5}{*}{\makecell{Radiomics\\Tibia}} 

& \makecell[l]{TC Original GLCM \\ Autocorrelation} & -1.004 & \makecell[c]{0.366 \\ (0.170, 0.720)} & 0.006\textsuperscript{\textbf{*}} & 1.559 \\
\cline{2-6}

& \makecell[l]{TC LoG First Order \\ 10\textsuperscript{th} Percentile} & -0.982 & \makecell[c]{0.375 \\ (0.181, 0.726)} & 0.005\textsuperscript{\textbf{*}} & 1.470 \\
\cline{2-6}

& \makecell[l]{TT Original First \\ Order Entropy} & -0.288 & \makecell[c]{0.750 \\ (0.320, 1.715)} & 0.497 & 1.482 \\
\cline{2-6}

& \makecell[l]{TT Exponential \\ GLCM IMC 1} & -1.230 & \makecell[c]{0.292 \\ (0.129, 0.601)} & 0.002\textsuperscript{\textbf{*}} & 2.041 \\
\cline{2-6}

& \makecell[l]{TT LBP First Order \\ Interquartile Range} & 0.813 & \makecell[c]{2.255 \\ (1.220, 4.506)} & 0.013\textsuperscript{\textbf{*}} & 1.499 \\ 

\hline
\multirow{5}{*}{\makecell{Radiomics\\Soft Tissue}}

& \makecell[l]{MT Gradient First \\ Order Maximum} & -1.225 & \makecell[c]{0.294 \\ (0.126, 0.600)} & 0.002\textsuperscript{\textbf{*}} & 1.627 \\
\cline{2-6}

& \makecell[l]{MT Logarithm First \\ Order Kurtosis} & -1.304 & \makecell[c]{0.271 \\ (0.112, 0.555)} & 0.001\textsuperscript{\textbf{*}} & 1.485 \\
\cline{2-6}

& \makecell[l]{AT Wavelet-H \\ GLCM IDMN} & -0.919 & \makecell[c]{0.399 \\ (0.187, 0.743)} & 0.008\textsuperscript{\textbf{*}} & 1.201 \\
\cline{2-6}

& \makecell[l]{AT Gradient \\ GLCM MCC} & 0.796 & \makecell[c]{2.217 \\ (1.244, 4.216)} & 0.010\textsuperscript{\textbf{*}} & 1.312 \\
\cline{2-6}

& \makecell[l]{AT Exponential First \\ Order Kurtosis} & -0.438 & \makecell[c]{0.645 \\ (0.297, 1.290)} & 0.236 & 1.217 \\   

\hline
\end{tabular}

\begin{tablenotes}
\footnotesize
\item[a] IMC: informational measure of correlation; IDMN: inverse difference moment normalized; MCC: maximal correlation coefficient 
\end{tablenotes}

\label{tab: classification model comparison stat (patient)}
\end{threeparttable}
\end{table}

In the non-radiomics model, the selected features captured key aspects of body composition, bone microarchitecture, density, and mechanical properties. More specifically, the remaining features included whole-body total fat mass percentage, trabecular number (TbN), cortical volumetric BMD (CtvBMD), bone stiffness, and the proximal trabecular-to-total bone failure load ratio ((Tb.F/TF)prox). Except for the last variable, all were significant (p $<$ 0.05) and inversely associated with osteoporosis, suggesting that increased adiposity, trabecular integrity, cortical density, and bone stiffness reduce the likelihood of osteoporosis.  

The radiomics tibia model incorporated two cortical and three trabecular features, of which four were statistically significant. As presented in Table \ref{tab: classification model comparison stat (patient)}, higher values of TC Original GLCM Autocorrelation and TC LoG First Order 10\textsuperscript{th} Percentile exhibit a negative association with osteoporosis, suggesting that greater cortical texture uniformity and higher cortical edge intensity reflect denser and structurally intact cortical bone. The TT Exponential GLCM IMC 1 is also linked to reduced osteoporosis risk, indicating that increased trabecular network complexity corresponds to healthier bones. Conversely, a higher TT LBP First Order Interquartile Range is associated with greater osteoporosis risk, reflecting increased local intensity variability and microstructural heterogeneity. 

The radiomics soft tissue model retained two significant features from the myotendinous region and two from the adipose region. As shown in Table \ref{tab: classification model comparison stat (patient)}, MT Gradient First Order Maximum and MT Logarithm First Order Kurtosis are both associated with lower odds of osteoporosis, suggesting that higher intensity gradients and greater peakedness in intensity distribution correspond to denser and healthier myotendinous tissue. Among the adipose features, AT Wavelet-H GLCM IDMN shows a negative association, indicating that higher local homogeneity in tissue texture is linked to lower osteoporosis risk, whereas AT Gradient GLCM MCC demonstrates a positive association, implying that increased texture complexity is related to greater osteoporosis susceptibility. 

\begin{table}[!t]
\centering
\caption{Classification performance of different models at the patient level on the held-out test set. Bold values indicate the best performance across models.}

\renewcommand{\arraystretch}{1.2}
\setlength{\tabcolsep}{4pt}
\begin{tabular}{lccccc}
\hline
\textbf{Model} & \textbf{Acc.} & \textbf{Sen.} & \textbf{Spe.} & \textbf{F1 Score} & \textbf{AUROC} \\ 

\hline

Non-Radiomics & 0.792 & 0.667 & \textbf{0.917} & 0.762 & 0.792 \\
Radiomics Tibia & 0.792 & 0.833 & 0.750 & 0.800 & 0.826 \\
Radiomics Soft Tissue & \textbf{0.875} & \textbf{0.917} & 0.833 & \textbf{0.880} & \textbf{0.875} \\

\hline
\end{tabular}
\label{tab: classification model comparison prediction (patient)}
\end{table}

As shown in Table \ref{tab: classification model comparison prediction (patient)}, both radiomics models outperformed the non-radiomics baseline in distinguishing osteoporosis from the control group in the held-out test set. In particular, the radiomics soft-tissue model achieved the highest accuracy (87.5\%), F1 score (0.88), and ROC-AUC (0.875), suggesting that radiomics features, particularly those extracted from soft-tissue compartments, may capture discriminative information that complements and extends beyond standard clinical and bone parameters. 

\section{Discussion}

The key findings from this study are as follows: 1) The proposed SegFormer-based segmentation network, which combines the representational strength of transformers with the efficiency of modern segmentation architectures, consistently outperformed conventional CNNs in both accuracy and generalization; 2) At the image level, radiomics features derived from the myotendinous and adipose tissue regions performed comparably to, or exceeded, the features extracted from bone compartments in predicting osteoporosis; 3) At the patient level, both radiomics-based models outperformed the non-radiomics baseline constructed from conventional variables; and 4) The radiomics soft tissue model achieved the highest overall accuracy, outperforming the radiomics bone model in osteoporosis classification at the patient level. 

Despite being the most common metabolic bone disease in the world, osteoporosis often remains undiagnosed until a fragility fracture occurs, underscoring the need for advanced computational tools to improve early detection \cite{Compston2019, Curtis2017}. Several studies have investigated the potential of radiomics for osteoporosis assessment using various imaging modalities, including radiography \cite{Kim2022, Fanelli2025}, CT \cite{Wang2023, Jiang2022, Huang2022, Xue2022, Tong2024}, QCT \cite{Xie2022}, and MRI \cite{He2021}. Extracted features have been used to construct radiomics score–based models \cite{Wang2023, Jiang2022} or as inputs to machine learning classifiers \cite{Huang2022, Xue2022}. Some studies further compared radiomics-based machine learning models with end-to-end CNNs for osteoporosis prediction \cite{Fanelli2025, Tong2024}. Wang \textit{et al.} \cite{Wang2023} and Xie \textit{et al.} \cite{Xie2022} integrated radiomics features with demographic and biochemical data, while Kim \textit{et al.} \cite{Kim2022} compared the predictive performance of clinical, radiomics, and deep learning features. However, prior works have largely focused on a single type of bone, such as the femur or vertebrae, without evaluating the distinct contributions of different bone compartments (such as cortical and trabecular) and surrounding soft tissues to osteoporosis classification. 

In this study, we develop a comprehensive machine learning framework to systematically investigate the discriminative information contained within distinct anatomical regions of HR-pQCT scans for osteoporosis prediction through radiomics analysis. Initially, a total of 6,720 individual HR-pQCT images are manually annotated to generate pixel-wise ground truth segmentation masks. A fully automated deep learning-based segmentation network, SegFormer, is implemented to delineate distal tibia HR-pQCT scans into five tissue classes, followed by a soft tissue segmentation protocol. Although HR-pQCT data are inherently volumetric, segmentation is performed in a 2D slice-wise manner to balance model complexity, computational feasibility, and training sample availability, improving model robustness. Following segmentation, radiomics features are extracted from each anatomical region, and multiple machine learning models are evaluated to identify their comparative predictive capabilities, forming a complete image level osteoporosis prediction pipeline. Finally, biological, clinical, physical function, and standard HR-pQCT parameters are incorporated into a multivariate non-radiomics logistic regression model to assess their predictive power relative to radiomics features at the patient level. 

The non-radiomics model confirms well-established associations, with higher trabecular number, greater cortical volumetric BMD, and increased bone stiffness more commonly observed in control subjects \cite{Agarwal2024, Okazaki2016}. A higher percentage of total body fat mass was also associated with the control group. However, prior literature has reported both positive and negative associations between fat mass and bone health, suggesting a highly complex and context-dependent relationship \cite{Zhao2008}. The radiomics soft-tissue model provides additional insight into these musculoskeletal relationships. The results suggest that denser myotendinous tissue and homogeneous adipose tissue may reflect healthier musculoskeletal composition, while complex or heterogeneous textural patterns may signal tissue alterations in osteoporosis. These observations are consistent with previous studies reporting that poor muscle quality, including increased fatty infiltration and reduced tissue density, is commonly observed in individuals with sarcopenia \cite{Warden2025}, a condition frequently associated with osteoporosis \cite{Laurent2019}. Similarly, findings from the radiomics tibia model indicate that dense, homogeneous cortical bone and complex trabecular microarchitecture are protective characteristics, whereas heterogeneous trabecular patterns indicate microstructural deterioration linked to osteoporosis. Overall, these findings demonstrate that radiomics features encode discriminatory information beyond standard clinical and density parameters and reveal that soft tissues may contain valuable predictive signals not captured by traditional bone-focused analyses. 

The proposed study has a few limitations. While the image level analysis utilized a relatively large dataset of 20,496 HR-pQCT images, the patient level classification was performed on a smaller cohort of 122 subjects. Moreover, the classification dataset was acquired from a single imaging center, which may limit generalizability. Future work should therefore include larger, multi-center datasets to validate the robustness of the proposed approach. In addition, although HR-pQCT offers unique advantages for resolving bone microarchitecture and soft-tissue compartments, its clinical availability remains limited compared with DXA, which may restrict near-term clinical translation. Nonetheless, insights from HR-pQCT-based radiomics may inform future algorithm development in more widely available and cost-effective imaging modalities. Finally, this study focused on the binary classification problem of osteoporosis versus control to establish a clear understanding of the distinct contributions of bone and soft tissue regions in HR-pQCT images. Subsequent research is needed to extend the framework to differentiate between osteoporosis, osteopenia, and normal bone mass for improved clinical applicability. 

\section*{Acknowledgment}

M. R. Subah thanks Amir Ali Dehghanpour for assistance with data preprocessing, and Farhan Sadik for manuscript review and constructive feedback throughout the study.

\section*{Conflict of Interest}

T. L. Nickolas reports consulting relationships with Amgen and serves on the Scientific Advisory Board of Pharmacosmos. All other authors declare no conflicts of interest related to this work.

\section*{References}

\bibliographystyle{ieeetr}
\bibliography{bibliography.bib}

\begin{thebibliography}{10}

\bibitem{Kanis2000}
J.~A. Kanis~et al., ``Long-term risk of osteoporotic fracture in malm\"{o},'' {\em Osteoporosis International}, vol.~11, p.~669–674, Sept. 2000.

\bibitem{Melton1998}
L.~J. Melton, E.~J. Atkinson, M.~K. O’Connor, W.~M. O’Fallon, and B.~L. Riggs, ``Bone density and fracture risk in men,'' {\em Journal of Bone and Mineral Research}, vol.~13, p.~1915–1923, Dec. 1998.

\bibitem{Compston2019}
J.~E. Compston, M.~R. McClung, and W.~D. Leslie, ``Osteoporosis,'' {\em The Lancet}, vol.~393, p.~364–376, Jan. 2019.

\bibitem{Curtis2017}
E.~M. Curtis, R.~J. Moon, N.~C. Harvey, and C.~Cooper, ``The impact of fragility fracture and approaches to osteoporosis risk assessment worldwide,'' {\em Bone}, vol.~104, p.~29–38, Nov. 2017.

\bibitem{Kanis2002}
J.~A. Kanis, ``Diagnosis of osteoporosis and assessment of fracture risk,'' {\em The Lancet}, vol.~359, p.~1929–1936, June 2002.

\bibitem{Bogl2010}
L.~H. Bogl, A.~Latvala, J.~Kaprio, O.~Sovij\"{a}rvi, A.~Rissanen, and K.~H. Pietil\"{a}inen, ``An investigation into the relationship between soft tissue body composition and bone mineral density in a young adult twin sample,'' {\em Journal of Bone and Mineral Research}, vol.~26, p.~79–87, Dec. 2010.

\bibitem{Reid2002}
I.~Reid, ``Relationships among body mass, its components, and bone,'' {\em Bone}, vol.~31, p.~547–555, Nov. 2002.

\bibitem{Osterhoff2016}
G.~Osterhoff, E.~F. Morgan, S.~J. Shefelbine, L.~Karim, L.~M. McNamara, and P.~Augat, ``Bone mechanical properties and changes with osteoporosis,'' {\em Injury}, vol.~47, p.~S11–S20, June 2016.

\bibitem{Warden2025}
S.~J. Warden~et al., ``Reference data and predictors of hr‐pqct‐derived muscle density and its prediction of physical performance,'' {\em Journal of Cachexia, Sarcopenia and Muscle}, vol.~16, July 2025.

\bibitem{Laurent2019}
M.~R. Laurent, L.~Dedeyne, J.~Dupont, B.~Mellaerts, M.~Dejaeger, and E.~Gielen, ``Age-related bone loss and sarcopenia in men,'' {\em Maturitas}, vol.~122, p.~51–56, Apr. 2019.

\bibitem{Kajiki2022}
Y.~Kajiki~et al., ``Psoas muscle index predicts osteoporosis and fracture risk in individuals with degenerative spinal disease,'' {\em Nutrition}, vol.~93, p.~111428, Jan. 2022.

\bibitem{Williams2021}
S.~Williams, L.~Khan, and A.~A. Licata, ``Dxa and clinical challenges of fracture risk assessment in primary care,'' {\em Cleveland Clinic Journal of Medicine}, vol.~88, p.~615–622, Nov. 2021.

\bibitem{Choksi2018}
P.~Choksi, K.~J. Jepsen, and G.~A. Clines, ``The challenges of diagnosing osteoporosis and the limitations of currently available tools,'' {\em Clinical Diabetes and Endocrinology}, vol.~4, May 2018.

\bibitem{Fuller2015}
H.~Fuller, R.~Fuller, and R.~M.~R. Pereira, ``High resolution peripheral quantitative computed tomography for the assessment of morphological and mechanical bone parameters,'' {\em Revista Brasileira de Reumatologia (English Edition)}, vol.~55, p.~352–362, July 2015.

\bibitem{Fink2018}
H.~A. Fink, L.~Langsetmo, T.~N. Vo, E.~S. Orwoll, J.~T. Schousboe, and K.~E. Ensrud, ``Association of high-resolution peripheral quantitative computed tomography (hr-pqct) bone microarchitectural parameters with previous clinical fracture in older men: The osteoporotic fractures in men (mros) study,'' {\em Bone}, vol.~113, p.~49–56, Aug. 2018.

\bibitem{SornayRendu2017}
E.~Sornay‐Rendu, S.~Boutroy, F.~Duboeuf, and R.~D. Chapurlat, ``Bone microarchitecture assessed by hr‐pqct as predictor of fracture risk in postmenopausal women: The ofely study,'' {\em Journal of Bone and Mineral Research}, vol.~32, p.~1243–1251, Mar. 2017.

\bibitem{Burt2018}
L.~A. Burt, S.~L. Manske, D.~A. Hanley, and S.~K. Boyd, ``Lower bone density, impaired microarchitecture, and strength predict future fragility fracture in postmenopausal women: 5-year follow-up of the calgary camos cohort,'' {\em Journal of Bone and Mineral Research}, vol.~33, p.~589–597, Jan. 2018.

\bibitem{Samelson2019}
E.~J. Samelson~et al., ``Cortical and trabecular bone microarchitecture as an independent predictor of incident fracture risk in older women and men in the bone microarchitecture international consortium (bomic): a prospective study,'' {\em The Lancet Diabetes \& amp; Endocrinology}, vol.~7, p.~34–43, Jan. 2019.

\bibitem{Agarwal2024}
S.~Agarwal~et al., ``Hr-pqct reveals marked trabecular and cortical structural deficits in women with pregnancy and lactation-associated osteoporosis (plo),'' {\em Journal of Bone and Mineral Research}, vol.~40, p.~38–49, Oct. 2024.

\bibitem{Okazaki2016}
N.~Okazaki, A.~J. Burghardt, K.~Chiba, A.~L. Schafer, and S.~Majumdar, ``Bone microstructure in men assessed by hr-pqct: Associations with risk factors and differences between men with normal, low, and osteoporosis-range areal bmd,'' {\em Bone Reports}, vol.~5, p.~312–319, Dec. 2016.

\bibitem{Cheng2024}
K.~Y.-K. Cheng~et al., ``Identification of osteosarcopenia by high-resolution peripheral quantitative computed tomography,'' {\em Journal of Personalized Medicine}, vol.~14, p.~935, Sept. 2024.

\bibitem{Mayerhoefer2020}
M.~E. Mayerhoefer~et al., ``Introduction to radiomics,'' {\em Journal of Nuclear Medicine}, vol.~61, p.~488–495, Feb. 2020.

\bibitem{Kim2022}
S.~Kim~et al., ``Deep radiomics–based approach to the diagnosis of osteoporosis using hip radiographs,'' {\em Radiology: Artificial Intelligence}, vol.~4, July 2022.

\bibitem{Fanelli2025}
F.~Fanelli~et al., ``Development of ai-based predictive models for osteoporosis diagnosis in postmenopausal women from panoramic radiographs,'' {\em Journal of Clinical Medicine}, vol.~14, p.~4462, June 2025.

\bibitem{Wang2023}
J.~Wang~et al., ``Prediction of osteoporosis using radiomics analysis derived from single source dual energy ct,'' {\em BMC Musculoskeletal Disorders}, vol.~24, Feb. 2023.

\bibitem{Jiang2022}
Y.-W. Jiang, X.-J. Xu, R.~Wang, and C.-M. Chen, ``Radiomics analysis based on lumbar spine ct to detect osteoporosis,'' {\em European Radiology}, vol.~32, p.~8019–8026, Apr. 2022.

\bibitem{Huang2022}
C.-B. Huang, J.-S. Hu, K.~Tan, W.~Zhang, T.-H. Xu, and L.~Yang, ``Application of machine learning model to predict osteoporosis based on abdominal computed tomography images of the psoas muscle: a retrospective study,'' {\em BMC Geriatrics}, vol.~22, Oct. 2022.

\bibitem{Xue2022}
Z.~Xue~et al., ``Using radiomic features of lumbar spine ct images to differentiate osteoporosis from normal bone density,'' {\em BMC Musculoskeletal Disorders}, vol.~23, Apr. 2022.

\bibitem{Tong2024}
X.~Tong, S.~Wang, J.~Zhang, Y.~Fan, Y.~Liu, and W.~Wei, ``Automatic osteoporosis screening system using radiomics and deep learning from low-dose chest ct images,'' {\em Bioengineering}, vol.~11, p.~50, Jan. 2024.

\bibitem{Xie2022}
Q.~Xie~et al., ``Development and validation of a machine learning-derived radiomics model for diagnosis of osteoporosis and osteopenia using quantitative computed tomography,'' {\em BMC Medical Imaging}, vol.~22, Aug. 2022.

\bibitem{He2021}
L.~He~et al., ``Radiomics based on lumbar spine magnetic resonance imaging to detect osteoporosis,'' {\em Academic Radiology}, vol.~28, p.~e165–e171, June 2021.

\bibitem{Valentinitsch2012}
A.~Valentinitsch~et al., ``Automated threshold-independent cortex segmentation by 3d-texture analysis of hr-pqct scans,'' {\em Bone}, vol.~51, p.~480–487, Sept. 2012.

\bibitem{Hafri2016}
M.~Hafri, R.~Jennane, E.~Lespessailles, and H.~Toumi, ``Dual active contours model for hr-pqct cortical bone segmentation,'' in {\em 2016 23rd International Conference on Pattern Recognition (ICPR)}, p.~2270–2275, IEEE, Dec. 2016.

\bibitem{Ohs2021}
N.~Ohs~et al., ``Automated segmentation of fractured distal radii by 3d geodesic active contouring of in vivo hr-pqct images,'' {\em Bone}, vol.~147, p.~115930, June 2021.

\bibitem{Klintstrm2024}
E.~Klintstr\"{o}m, B.~Klintstr\"{o}m, O.~Smedby, and R.~Moreno, ``Automated region growing-based segmentation for trabecular bone structure in fresh-frozen human wrist specimens,'' {\em BMC Medical Imaging}, vol.~24, May 2024.

\bibitem{Zhou2025}
M.~Zhou~et al., ``The impact of segmentation approach on hr-pqct microarchitectural and biomechanical metrics depends on bone structure,'' {\em Journal of Bone and Mineral Research}, vol.~40, p.~868–880, Apr. 2025.

\bibitem{Burghardt2010}
A.~J. Burghardt, H.~R. Buie, A.~Laib, S.~Majumdar, and S.~K. Boyd, ``Reproducibility of direct quantitative measures of cortical bone microarchitecture of the distal radius and tibia by hr-pqct,'' {\em Bone}, vol.~47, p.~519–528, Sept. 2010.

\bibitem{Whittier2020}
D.~E. Whittier~et al., ``Guidelines for the assessment of bone density and microarchitecture in vivo using high-resolution peripheral quantitative computed tomography,'' {\em Osteoporosis International}, vol.~31, p.~1607–1627, May 2020.

\bibitem{Neeteson2023}
N.~J. Neeteson, B.~A. Besler, D.~E. Whittier, and S.~K. Boyd, ``Automatic segmentation of trabecular and cortical compartments in hr-pqct images using an embedding-predicting u-net and morphological post-processing,'' {\em Scientific Reports}, vol.~13, Jan. 2023.

\bibitem{Ronneberger2015}
O.~Ronneberger, P.~Fischer, and T.~Brox, {\em U-Net: Convolutional Networks for Biomedical Image Segmentation}, p.~234–241.
\newblock Springer International Publishing, 2015.

\bibitem{ViT}
A.~Dosovitskiy~et al., ``An image is worth 16x16 words: Transformers for image recognition at scale,'' in {\em International Conference on Learning Representations (ICLR)}, 2021.

\bibitem{SegFormer}
E.~Xie, W.~Wang, Z.~Yu, A.~Anandkumar, J.~M. Alvarez, and P.~Luo, ``Segformer: simple and efficient design for semantic segmentation with transformers,'' in {\em Proceedings of the 35th International Conference on Neural Information Processing Systems}, NIPS '21, Curran Associates Inc., 2021.

\bibitem{Subah2024}
M.~R. Subah, M.~A.~G. Zilani, A.~A. Dehghanpour, and R.~K. Surowiec, ``A feature-based fracture risk prediction model from hr-pqct imaging,'' {\em Journal of Bone and Mineral Research}, vol.~39, pp.~i345--i345, Nov. 2024.

\bibitem{Subah2025}
M.~R. Subah, M.~A.~G. Zilani, T.~L. Nickolas, M.~R. Allen, S.~J. Warden, and R.~K. Surowiec, ``Automated segmentation and radiomic analysis of soft tissue features distinguishes high fragility risk in chronic kidney disease using hrpqct,'' {\em Journal of Bone and Mineral Research}, vol.~40, pp.~i168--i168, Nov. 2025.

\bibitem{Warden2021}
S.~J. Warden, Z.~Liu, R.~K. Fuchs, B.~van Rietbergen, and S.~M. Moe, ``Reference data and calculators for second-generation hr-pqct measures of the radius and tibia at anatomically standardized regions in white adults,'' {\em Osteoporosis International}, vol.~33, p.~791–806, Sept. 2021.

\bibitem{Sode2011}
M.~Sode, A.~J. Burghardt, J.-B. Pialat, T.~M. Link, and S.~Majumdar, ``Quantitative characterization of subject motion in hr-pqct images of the distal radius and tibia,'' {\em Bone}, vol.~48, p.~1291–1297, June 2011.

\bibitem{Fedorov2012}
A.~Fedorov~et al., ``3d slicer as an image computing platform for the quantitative imaging network,'' {\em Magnetic Resonance Imaging}, vol.~30, p.~1323–1341, Nov. 2012.

\bibitem{Cordts2016}
M.~Cordts~et al., ``The cityscapes dataset for semantic urban scene understanding,'' in {\em 2016 IEEE Conference on Computer Vision and Pattern Recognition (CVPR)}, p.~3213–3223, IEEE, June 2016.

\bibitem{AdamOptim}
D.~P. Kingma and J.~Ba, ``Adam: A method for stochastic optimization,'' in {\em International Conference on Learning Representations (ICLR)}, 2015.

\bibitem{Erlandson2017}
M.~C. Erlandson~et al., ``Muscle and myotendinous tissue properties at the distal tibia as assessed by high-resolution peripheral quantitative computed tomography,'' {\em Journal of Clinical Densitometry}, vol.~20, p.~226–232, Apr. 2017.

\bibitem{Hildebrand2021}
K.~N. Hildebrand, K.~Sidhu, L.~Gabel, B.~A. Besler, L.~A. Burt, and S.~K. Boyd, ``The assessment of skeletal muscle and cortical bone by second-generation hr-pqct at the tibial midshaft,'' {\em Journal of Clinical Densitometry}, vol.~24, p.~465–473, July 2021.

\bibitem{vanGriethuysen2017}
J.~J.~M. van Griethuysen~et al., ``Computational radiomics system to decode the radiographic phenotype,'' {\em Cancer Research}, vol.~77, p.~e104–e107, Oct. 2017.

\bibitem{Cella2007}
D.~Cella~et al., ``The patient-reported outcomes measurement information system (promis): Progress of an nih roadmap cooperative group during its first two years,'' {\em Medical Care}, vol.~45, p.~S3–S11, May 2007.

\bibitem{oktay2018attention}
O.~Oktay~et al., ``Attention u-net: Learning where to look for the pancreas,'' in {\em Medical Imaging with Deep Learning}, 2018.

\bibitem{Haralick1973}
R.~M. Haralick, K.~Shanmugam, and I.~Dinstein, ``Textural features for image classification,'' {\em IEEE Transactions on Systems, Man, and Cybernetics}, vol.~SMC-3, p.~610–621, Nov. 1973.

\bibitem{Zhao2008}
L.-J. Zhao~et al., ``Correlation of obesity and osteoporosis: Effect of fat mass on the determination of osteoporosis,'' {\em Journal of Bone and Mineral Research}, vol.~23, p.~17–29, Jan. 2008.

\end{thebibliography}

\end{document}